# Inserting Faces inside Captions: Image Captioning with Attention Guided Merging

Yannis Tevissen[1], Khalil Guetari[1], Marine Tassel[1],
Erwan Kerleroux[1] and Frédéric Petitpont[1]

[1] Moments Lab Research, 92012 Boulogne-Billancourt, France
`yannis.tevissen@momentslab.com`

**Abstract.** Image captioning models are widely used to describe recent and archived pictures with the objective of improving their accessibility and retrieval. Yet, these approaches tend to be inefficient and biased at retrieving people's names. In this work we introduce AstroCaptions, a dataset for the image captioning task. This dataset specifically contains thousands of public figures that are complex to identify for a traditional model. We also propose a novel post-processing method to insert identified people's names inside the caption using explainable AI tools and the grounding capabilities of vision-language models. The results obtained with this method show significant improvements of captions quality and a potential of reducing hallucinations. Up to 93.2% of the persons detected can be inserted in the image captions leading to improvements in the BLEU, ROUGE, CIDEr and METEOR scores of each captioning model.

**Keywords:** Image Captioning, Face Identification, Visual Grounding.

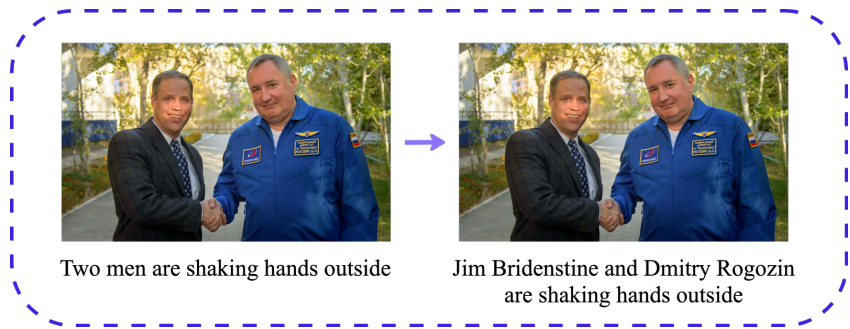

**Fig. 1.** Example of the results obtained with the presented method. People's names are inserted inside the caption.

## 1    Introduction

Media rights holders gather several billion images that together represent a big part of the world's cultural heritage. However, most of the time these archives are stored unstructured and without the metadata needed for its usage. Even more images are shared every day on social networks but again, most of them are not accompanied by a caption preventing also their accessibility and retrieval. Yet captions are a key metadata for archivists to improve the indexing and therefore usability and retrieval of the multimedia contents they manage.

Image captioning is the task of automatically describing an image with a generated text. Most of the approaches rely on Vision Language Models (VLMs) and form together a rich state-of-the-art [1]. The best methods are often large pretrained models finetuned on the Microsoft COCO dataset [2] that provides a large variety of objects and scenes together with the associated caption written by annotators. This dataset, like many others, has the issue of being stuck in time. MS-COCO for instance was released in 2015 and got only a few upgrades since. Even if this is not the philosophy behind very generic datasets, this clearly would not be sufficient to integrate in such a dataset every public figure that was either relevant today or back in the days. More generally, one cannot expect an image captioning model to perform efficient person identification at scale without giving it a significant number of images with every person during the training phase. On the other hand, face identification is a very mature technology and only a few images of a face are now necessary to make it identifiable even among thousands of others [3], [4].

However, when the image is about one or several public figures, the name of this person is a key element of the caption. To obtain captions with accurate person names included, the only option was to train or finetune a captioning model on a very specific dataset. Indeed, when one tries to obtain domain specific captions, one can already train a dedicated model with structured or unstructured data and obtain detailed results even on very specific objects [5].

In this work, we propose a novel post-processing method, that requires no training, to insert identified people's names inside image captions. This method takes advantages of the emergent grounding capabilities of VLMs. We also introduce AstroCaptions, a novel domain-specific image captioning dataset with a high number of recognizable public figures.

### 1.1 Related works

Several methods have already been used to improve image captioning and large language models results. LLMs have been used to fuse and rank captions [6]. With WikiChat [7], few-shots grounding on Wikipedia allowed for a reduction of hallucinations and up to 97.3% factual accuracy in simulated conversations with popular LLMs.

Recently, a lot of interest has also been given to the grounding capabilities of VLMs. This makes their result more interpretable and allows for further integration in complex pipelines [8], [9].

Finally, explainability of LLMs and VLMs has also been a rising topic in AI with numerous methods attempting to mitigate the black-box effect of AI models by increasing the results interpretability [10]. These methods can help a human reader to visualize the computation steps that lead to a deep neural network output. One of the most used visualization methods are attention maps that visually convey on an image the importance given to each pixel to produce a certain result. For a certain image captioning model, one can visualize for each word what were the parts of the image that are linked to it.

## 2  Archive dataset

Archive datasets are very challenging as they are often domain- specific. They carry domain-related imagery and people who often don't appear in sufficient quantity in the data used for large model pretraining. Therefore, performances tend to be very low for archive image captioning.

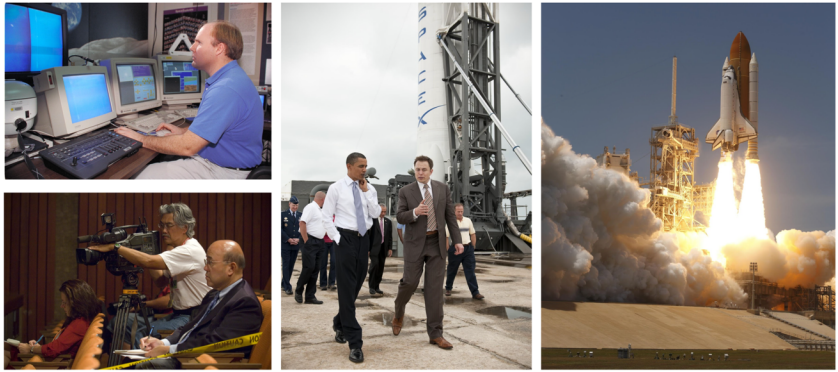

**Fig. 2.** Example images taken from the AstroCaptions dataset.

For this study, we introduce AstroCaptions, a novel dataset made of 44115 publicly available NASA archive images. It contains both very recent photos and old archive pictures from the first Apollo missions. Many astronauts, NASA scientists and executives appear on these images. Each image comes with a description, scraped from public NASA website. These provides both visual description of the image and contextual information. NLTK [11] is used to split the descriptions to keep only the first sentence, often containing the factual descriptions. Given the unstructured nature of the data, this part is particularly challenging, and the final dataset still contains some dates and event names that could not be inferred by a captioning model. That is why, with a few-shot approach, we also used OpenAI GPT-4 model to create new captions from the long descriptions. An example of all the captions available in the dataset for one image is shown in Table 1.

**Table 1.** Example of the different captions available in the AstroCaptions dataset.

| NASA image description | First sentence split | GPT-4 caption |
|---|---|---|
| Dr. Donald Gilles, the Discipline Scientist for Materials Science in NASA's Microgravity Materials Science and Applications Department, demonstrates to Carl Dohrman a model of dendrites, the branch-like structures found in many metals and alloys. Dohrman was recently selected by the American Society for Metals International as their 1999 ASM International Foundation National Merit Scholar. The University of Illinois at Urbana-Champaign freshman recently toured NASA's materials science facilities at the Marshall Space Flight Center. | Dr. Donald Gilles, the Discipline Scientist for Materials Science in NASA's Microgravity Materials Science and Applications Department, demonstrates to Carl Dohrman a model of dendrites, the branch-like structures found in many metals and alloys. | Dr. Donald Gilles showing Carl Dohrman a model of dendritic structures in a laboratory setting. |

The names of the people appearing in the dataset are also extracted from the captions using a finetuned BERT model [12] for the task of named entity recognition. A total of 13083 identifiable persons appears on the created dataset. For this study, a small database is also created with the 100 most appearing people and their faces. Thanks to this, we were able to identify a significant amount of people within the images, so we added to this dataset the faces that we detected and identified during our experiments using a state-of-the-art face detection and identification system. The number of identified people per image is very heterogenous as depicted in the distribution on Fig. 3.

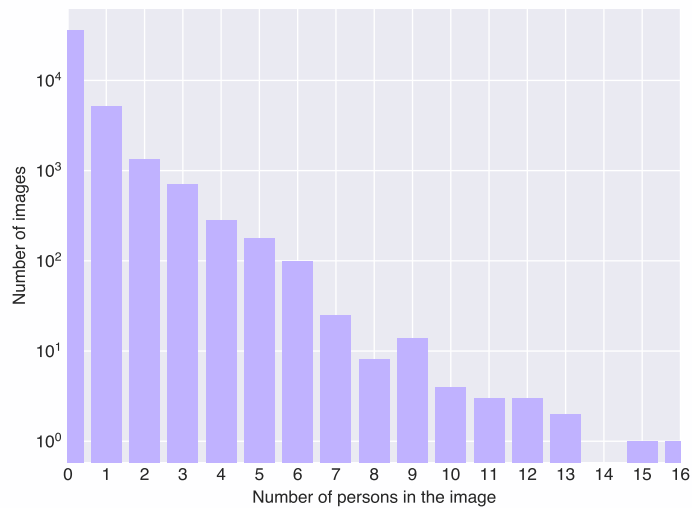

**Fig. 3.** Distribution of the number of identified people on the images of the AstroCaptions dataset.

## 3   Experiment

### 3.1   Image captioning baseline

The method we propose requires a base model to generate an unmerged caption. Although this method is relatively agnostic to the base model used, for this work we choose to use state-of-the-art captioning models such as BLIP 2 [13], Instruct-BLIP [14] and LLaVa-1.6 [15]. The BLIP architecture, firstly introduced in [16] is composed of a CLIP features extraction module [17], a frozen LLM and a QFormer network trained to pass from one embedding to the other. For this study BLIP2 was used with a FlanT5-XXL model [18]. InstructBLIP follows the same architecture with a finetuned QFormer and, in our case, a Vicuna-7B LLM [19]. Finally, LLaVa-1.6, the last generation of the LLaVa architecture based models [20], [21], is also used with a Mistral 7B [22] for text generation. All the models were run using float 16 precision.

### 3.2 Face detection and identification

For each image, faces are detected using YuNet [23], a lightweight model designed to run fast inferences at scale and on CPU. Because of its simplicity of use, the AWS Rekognition API[1] is used to host our thesaurus of the 100 most appearing people. For each of the detected faces, the faces similarities between these and the database is computed by AWS online service and a similarity score is obtained. For the dataset and our merging strategy, we only keep faces that are above 90% of confidence during the identification phase.

### 3.3 Attention guided fusion

To perform the insertion of person names in the caption, several candidate words such as "man", "woman", "person", "astronaut" were identified. For each candidate word in the predicted caption, an attention map is generated using OmniXAI [24] and a BLIP model finetuned for the image captioning task on the COCO dataset. Thanks to the emergent grounding capabilities of this model, the generated heatmap highlights the parts of the image that the model links to the word.

This heatmap extraction was run using two different sizes of the BLIP model. In the following section base size merging refers as the use of BLIP trained with a base vision transformer and the Captioning and Filtering strategy [16] whereas large merging refers as the use of the BLIP with a large vision transformer encoder. Both models where pretrained on 129 million images.

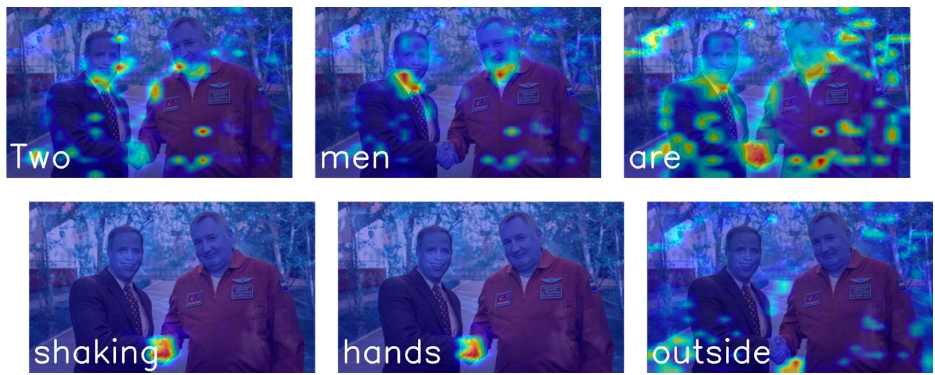

**Fig. 4.** Attention map generated with OmniXAI and the base size BLIP model with reference caption "Two men are shaking hand outside".

Once the attention maps obtained for each candidate word, the overlap percentage between the activated areas and the bounding box of each identified faces faces are computed. If it is above a certain threshold $\Theta$, the merging is applied. For this study, we found empirical optimum with $\Theta = 0.05$.

---

[1] https://docs.aws.amazon.com/rekognition/

Several merging rules have also been written to ensure the correctness of the syntax of the new caption. In the example presented in Fig. 4, the word "two" needs to be removed only if two faces are available to be inserted.

Three of the most frequent merging scenarios are illustrated in the Table 2. The first one constitutes the simplest case where one candidate word is replaced by one name. The second example is a case where the grounding capacities of the VLM are useful to replace the correct candidate word in the base caption. Finally, the third example is an example where the proposed method corrects the name wrongly suggested by the image captioning model.

**Table 2.** Example of outputs generated by the presented method.

| Image | Base caption | Enhanced caption |
|---|---|---|
| 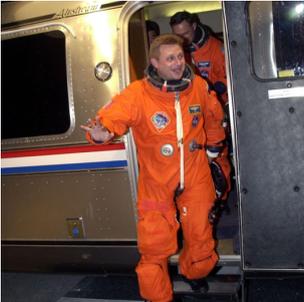 | a man in an orange space suit | Yuri Onufrienko in an orange space suit |
| 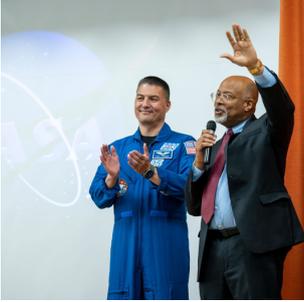 | An astronaut clapping with a man in a suit | Kjell Lindgren clapping with Glenn Ivey in a suit |
| 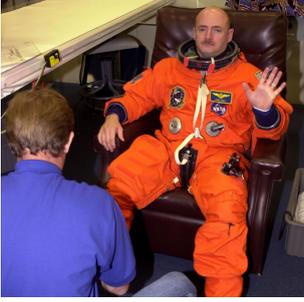 | Chris Hadfield in a space suit | Mark E. Kelly in a space suit |

**LLMs for complex cases.** Depending on the structure of the caption, some candidate words are trickier to replace than others. For the most complex cases, a Falcon-40B [25] large language model can be prompted to insert the identified people's names at the right place in the sentence. We remove these cases from the presented results as they introduce a lot of new biases and dependencies to the LLM chosen.

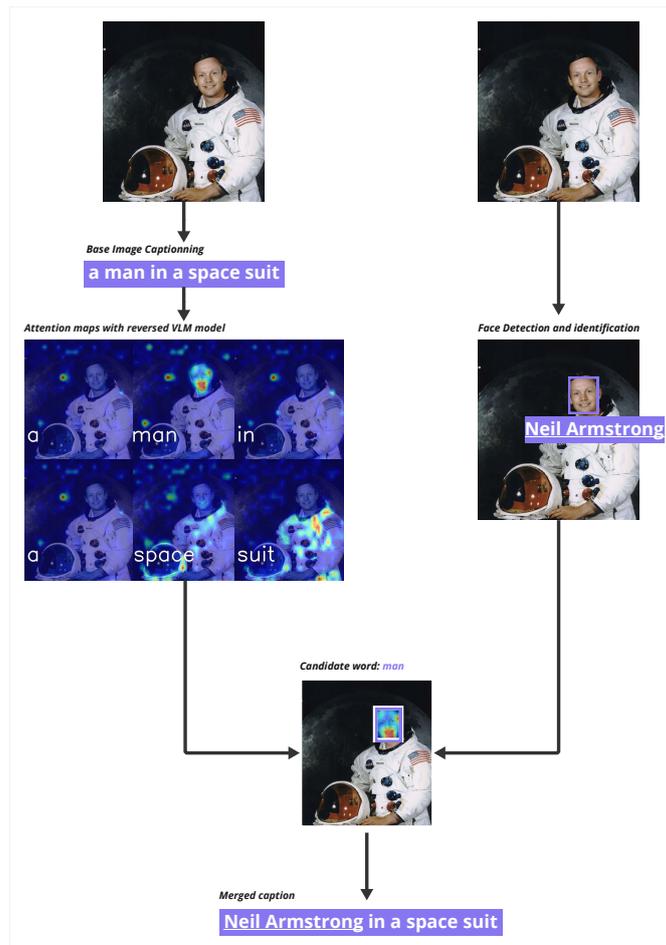

**Fig. 5.** Overview of the method presented to insert people's names in the generated captions.

Finally, after the whole analysis and post-processing steps described in Fig. 5, a new caption is obtained with the identified people names inserted.

## 4   Results

The results are reported using the following metrics. BLEU [26], ROUGE [27] are standard metrics for text generation, CIDEr [28] and METEOR [29] are specialized image captioning evaluation metrics. These compute in various ways a semantic overlap between the ground truth caption and the inferred one. Additionally, in Table 3, we report the percentages of identified persons that are successfully inserted in the captions.

### 4.1   Names insertion

**Table 3.** Results of the face identification and insertion phase of the presented method on the AstroCaptions dataset.

| Model | Number of person inserted | % of person detected |
| --- | --- | --- |
| BLIP2-FlanT5-XXL + base merging | 10 608 | 81.08 |
| BLIP2-FlanT5-XXL + large merging | 9414 | 71.95 |
| InstructBLIP-Vicuna-7B + base merging | **12193** | **93.20** |
| InstructBLIP-Vicuna-7B + large merging | 12181 | 93.11 |
| LLaVA-1.6-Mistral 7B + base merging | 10907 | 83.37 |
| LLaVA-1.6-Mistral 7B + large merging | 10840 | 82.86 |

Among the 44115 images of the dataset, 13083 persons were identified over 7820 unique images and, depending on the model and the merging strategy, up to 12193 were inserted in the caption. As shown in Table 3 the large-scale merging strategy leads to a lower number of people's names inserted in the captions.

### 4.2   Caption quality measure

As the results of some metrics were very low when using the NASA ground truth, we report our results in two parts. Table 4 show the exploitable results of the predicted captions when compared with the one-sentence ground truth captions whereas Table 5 show the extensive results with GPT-4 captions used as reference.

The first results show a significant improvement of the BLEU metric with up to 87.5% raise of the BLIP2 score with the merging strategy, 34,8% improvement for InstructBLIP and 11.8% for LLaVa.

Table 4. Results obtained with the one-sentence ground truth caption.

| Method | # of params | BLEU | ROUGE |
|---|---|---|---|
| BLIP2-FlanT5-XXL | 12.1B | 0.48 | 0.06 |
| InstructBLIP-Vicuna-7B | 7.2B | 0.46 | 0.06 |
| LLaVA-1.6-Mistral 7B | 7.6B | 1.19 | 0.07 |
| BLIP2-FlanT5-XXL + base merging | - | 0.63 | 0.06 |
| BLIP2-FlanT5-XXL + large merging | - | 0.9 | **0.08** |
| InstructBLIP-Vicuna-7B + base merging | - | 0.61 | 0.07 |
| InstructBLIP-Vicuna-7B + large merging | - | 0.62 | 0.07 |
| LLaVA-1.6-Mistral 7B + base merging | - | 1.19 | 0.7 |
| LLaVA-1.6-Mistral 7B + large merging | - | **1.33** | 0.7 |

The results obtained when we compare the predicted captions with GPT-4 are significantly higher as the reference captions are closer to what an image captioning model can generate. Except for LLaVa and its already high BLEU metric, the fusion strategy helps improving the results of every model on every metric with up to 5.08 points gained on CIDEr with BLIP and 3.45 points with InstructBLIP.

InstructBLIP together with the large merging strategy performs particularly well in terms of ROUGE and CIDEr but LLaVa seems more stable across all the computed metrics.

Table 5. Results obtained with the GPT-4 generated caption used as reference.

| Method | BLEU | ROUGE | CIDEr | METEOR |
|---|---|---|---|---|
| BLIP2-FlanT5-XXL | 10.40 | 14.84 | 12.70 | 4.78 |
| InstructBLIP-Vicuna-7B | 10.92 | 15.46 | 14.25 | 4.95 |
| LLaVA-1.6-Mistral 7B | **16.81** | 15.22 | 14.45 | 6.11 |
| BLIP2-FlanT5-XXL + base merging | 13.46 | 15.34 | 15.65 | 5.54 |
| BLIP2-FlanT5-XXL + large merging | 16.35 | 15.71 | 17.78 | 6.01 |
| InstructBLIP-Vicuna-7B + base merging | 14.31 | 16.06 | 17.83 | 5.86 |
| InstructBLIP-Vicuna-7B + large merging | 14.32 | **16.07** | **17.90** | 5.86 |
| LLaVA-1.6-Mistral 7B + base merging | 16.38 | 15.61 | 16.71 | **6.85** |
| LLaVA-1.6-Mistral 7B + large merging | 16.38 | 15.60 | 16.70 | 6.84 |

## 5    Discussion

As expected, we obtain better results with LLaVa as it is the most recent and advanced image captioning model among the three that were tested. We also notice that the merging strategy improves the results of every model.

The difference between the different models' behaviors can be explained by their variable grounding capabilities. For instance, InstructBLIP is known to better ground objects than BLIP2 [30] which can explain why more people are inserted in InstructBLIP captions.

Although the method shows significant improvements in the BLEU score, the overall results are quite low, when the predicted captions are compared with the ground truth NASA captions. this can be explained by the fact that some captions are more contextual than descriptive.

### 5.1    Energetic impact

The method presented allows for a better captioning of images without the need of any new pretraining or finetuning. This makes it one of the most efficient and cost-effective methods for large captioning jobs requiring accurate person identification. This post processing also uses relatively small models both for face detection and heatmap generation. The whole process can easily be run on CPU making it also less computationally heavy than methods that would need larger pretrained models.

### 5.2    Limits of the method

This method works only on faces and requires a sufficiently large and topical face dataset to demonstrate its effectiveness. This can be achieved with relatively low effort by using off-the-shelf face identification APIs such as AWS Rekognition. As for any technology using biometric, we recommend using it with extreme attention given to the privacy and consent of the people being identified.

### 5.3    Societal impact

We also want to highlight that, if merging with faces can help reduce biases, this can also introduce new types of biases such as the ones that can be embedded in the face detection and identification used [31].

On top of that, inserting people's names inside captions helps improve the relevance of these texts but it also allows for a new way to link people with the actions they are performing or the places they are seen in. Such merging and outputs should be treated with caution and with the consent of the person whose face is being identified and then merged.

### 5.4    Reproducible research

All the models used in this work have open-sourced weights. Concerning the data, we make our work fully reproducible by sharing on HuggingFace the AstroCaptions dataset introduced for this study: https://huggingface.co/datasets/momentslab/AstroCaptions

### 5.5    Future works

We believe these results can be reproduced with many other expert systems such as object and landmark detection and identification. Implementing several of these can reduce the need of training bigger captioning models while also improving the overall precision and factuality of their captions.

Recent works also demonstrate that BLIP models are not the best models at grounding tasks. Yet this is what most of this method relies on. Replacing the BLIP model used for image-text matching by a model specialized for grounding such as GEM [30] could also lead to further improvements of the proposed method.

## 6      Conclusion

In this study, we introduced a new dataset called AstroCaptions for the task of image captioning. A new post-processing method for inserting people's names inside captions has also demonstrated promising results. Indeed, the results presented show that the post-processing strategy helps improving the quality of the captions by inserting a significant amount of the identified people directly in the captions. This work highlights the complementarity of large vision-language models and smaller expert systems for the accurate description of domain specific images.